\documentclass[10pt,twocolumn,letterpaper]{article}

\usepackage{cvpr}
\ifdefined\pdfminorversion\pdfminorversion=7\fi
\ifdefined\pdfsuppressptexinfo\pdfsuppressptexinfo=-1\relax\fi

\usepackage{wrapfig}
\usepackage{colortbl}
\usepackage{arydshln}
\expandafter\let\csname endtabular*\endcsname\endtabular
\usepackage{multirow}
\usepackage{pifont}

\definecolor{graytext}{gray}{0.3}
\definecolor{bestresultbg}{HTML}{D7EEE5}
\definecolor{secondresultbg}{HTML}{E9EFFB}
\newcommand{\cmark}{\textcolor{green!60!black}{\ding{51}}}
\newcommand{\xmark}{\textcolor{red}{\ding{55}}}
\newcommand{\bestresult}[1]{\cellcolor{bestresultbg}\textbf{#1}}
\newcommand{\secondresult}[1]{\cellcolor{secondresultbg}\underline{#1}}
\DeclareRobustCommand{\bestlegend}[1]{{\setlength{\fboxsep}{1pt}\colorbox{bestresultbg}{\textbf{#1}}}}
\DeclareRobustCommand{\secondlegend}[1]{{\setlength{\fboxsep}{1pt}\colorbox{secondresultbg}{\underline{#1}}}}
\newcommand\MPVPE{MPVPE$\downarrow$}
\newcommand\MPJPE{MPJPE$\downarrow$}
\newcommand\POEMs{POEM$_{\text{v1}}$\xspace}
\newcommand\POEMg{POEM$_{\text{v2}}$\xspace}

\usepackage{placeins}
\definecolor{cvprblue}{rgb}{0.21,0.49,0.74}
\usepackage[pagebackref,breaklinks,colorlinks,allcolors=cvprblue]{hyperref}
\hypersetup{
  pdftitle={FReSH: Feed-Forward Joint Reconstruction of Scene-Hand from Uncalibrated Sparse Videos},
  pdfauthor={Yumeng Liu, Xiao-Xiao Long, Marc Habermann, Xuanze Yang, Cheng Lin, Yuan Liu, Yuexin Ma, and Ligang Liu}
}

\title{FReSH: Feed-Forward Joint Reconstruction of Scene-Hand from Uncalibrated Sparse Videos}

\author{
Yumeng Liu$^{1}$\quad Xiao-Xiao Long$^{2}$\quad Marc Habermann$^{3}$\quad Xuanze Yang$^{1}$\\
Cheng Lin$^{4}$\quad Yuan Liu$^{5}$\quad Yuexin Ma$^{6}$\quad Ligang Liu$^{1}$\thanks{Corresponding author.}\\[0.5em]
{\small $^{1}$University of Science and Technology of China}\quad
{\small $^{2}$Nanjing University}\\
{\small $^{3}$Max-Planck-Institut f\"ur Informatik}\quad
{\small $^{4}$Macau University of Science and Technology}\\
{\small $^{5}$Hong Kong University of Science and Technology}\quad
{\small $^{6}$ShanghaiTech University}\\[0.35em]
{\small\textcolor{cvprblue}{\textbf{Project Page:}}
\href{https://lym29.github.io/FReSH/}{\texttt{https://lym29.github.io/FReSH/}}}
}

\makeatletter
\apptocmd{\@maketitle}{%
  \par\noindent
  \begin{minipage}{\textwidth}
    \centering
    \includegraphics[width=\linewidth]{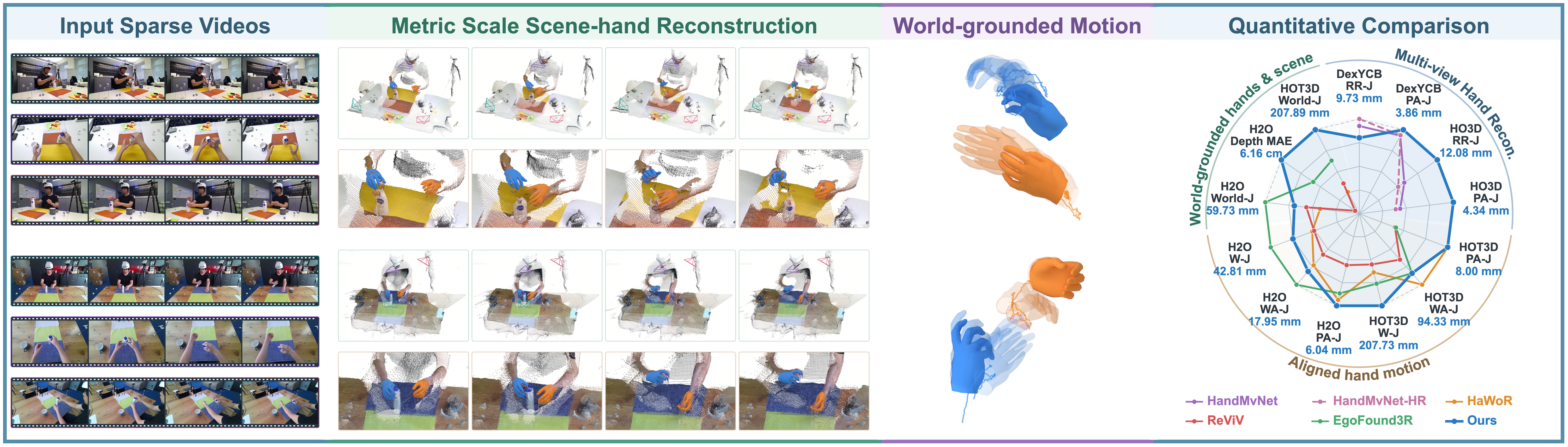}
    \captionsetup{hypcap=false}
    \captionof{figure}{\textbf{FReSH} turns sparse, uncalibrated videos into metric-scale scenes and world-grounded hand motion with a feed-forward model.}
    \label{fig:teaser}
  \end{minipage}
  \par\vspace{1em}
}{}{\PackageError{FReSH}{Could not append the first-page teaser}{}}
\makeatother

\begin{document}
\raggedbottom
\maketitle

\begin{abstract}

Reconstructing hand motion and scene geometry from video in a shared world frame provides valuable information about how people perform manipulation tasks in their surroundings.
Collecting such 3D data has often relied on dense arrays of calibrated cameras, restricting where recordings can be made and making capture difficult to scale.
Sparse-view video from one or a few uncalibrated RGB cameras simplifies capture, but occlusion, depth ambiguity, and unknown camera poses complicate reconstruction in a shared world frame.
To address these limitations, we present FReSH, a feed-forward framework that jointly reconstructs hand motion, scene geometry, and camera trajectories from videos captured by a sparse set of uncalibrated cameras.
FReSH makes scene geometry and hand articulation mutually informative: bidirectional feature exchange lets global context guide hand reconstruction and fine hand details inform scene estimation. Hand joints serve as anatomical correspondences for fusing evidence across views, supporting hand and camera estimation. In temporal space, a bounded causal memory preserves hand and scene context over time, while learned registration anchors successive reconstructions to a fixed world frame.
Extensive experiments on established benchmarks demonstrate that FReSH achieves high-quality reconstruction of both hands and scenes. With uncalibrated multi-view images as input, FReSH even surpasses calibrated methods in hand vertex reconstruction accuracy, while also achieving competitive joint hand--scene reconstruction from sparse, uncalibrated videos.
\end{abstract}

\section{Introduction}
\label{sec:intro}

From grasping objects to using tools, humans rely on coordinated hand movements to interact with the physical world. Reconstructing these movements in 3D supports applications in human behavior analysis, virtual reality, and embodied AI, to only name a few.~\cite{qi2024computer,pei2022hand,grauman2022ego4d,wang2025dexh2r}. For the development of these fields, human manipulation demonstrations with accurate annotations of hand motion and scene geometry are especially valuable, as they provide supervision for learning how actions relate to the physical environment~\cite{wang2024dexcap,qin2022dexmv}. However, acquiring such annotations has often relied on carefully calibrated multi-camera rigs~\cite{chao2021dexycb,kwon2021h2o}, tying data collection to dedicated capture environments. We therefore seek to lower this barrier and support scalable data collection with ordinary and uncalibrated phone cameras. Specifically, we study joint reconstruction of hand motion, cameras, and scene geometry in a shared world frame from sparse uncalibrated videos. Achieving this is challenging: sparse views and occlusion leave observations incomplete, while unknown camera geometry makes it difficult to establish a common spatial reference and disentangle hand motion from camera motion~\cite{zhang2025hawor}.

3D hand reconstruction has progressed from estimating articulation in individual images~\cite{pavlakos2024reconstructing,potamias2025wilor} to combining observations across views~\cite{yang2024POEM,ali2025handmvnet}. These methods focus on hand geometry, with camera and scene reconstruction outside their scope. Staged video pipelines estimate hand and camera motion separately and combine them in world coordinates~\cite{zhang2025hawor}. Iterative optimization adds computational cost, and errors in camera estimates can carry over to the recovered hand motion. Recent feed-forward approaches jointly estimate hand and camera motion~\cite{zhu2026mint} or recover hands alongside scene geometry~\cite{lyu2026reviv,fu2026egofound3r}. These methods target monocular egocentric video, leaving open how to integrate evidence across sparse, uncalibrated views and maintain a shared world frame over time.

In parallel, the broader 3D vision community has witnessed a paradigm shift towards feed-forward 3D reconstruction, exemplified by DUSt3R~\cite{dust3r_cvpr24}, MASt3R~\cite{mast3r_eccv24}, and VGGT~\cite{wang2025vggt,wang2026vggtomega}. These foundation models demonstrate that scene geometry can be learned directly from large-scale data, with VGGT jointly predicting camera parameters, depth, and point tracks through a unified transformer. Recent works such as Human3R~\cite{chen2025human3r} and UniSH~\cite{li2026unish} further explore joint human-scene reconstruction. Nevertheless, directly applying scene-level models to hand reconstruction remains challenging. Full-image inputs may provide limited information about fine articulation when hands occupy a small fraction of the frame, while tight hand crops preserve hand detail but sacrifice scene context and make camera pose estimation harder. This motivates a more integrated approach that jointly reasons about hand geometry and camera pose, using both detailed hand observations and the broader scene context.

To bridge this gap, we introduce \textbf{FReSH}, a feed-forward framework for joint hand, scene, and camera reconstruction from uncalibrated multi-view observations. Our central insight is that hand articulation and scene geometry provide complementary cues for establishing a common spatial reference across views. We combine hand and scene features in a dual-branch architecture for spatial reconstruction. The global branch encodes full images to capture scene structure and geometric relationships across views, while the local branch encodes hand crops to retain the detail needed to recover articulation. The two branches exchange features through bidirectional cross-attention. FReSH uses predicted 2D hand joints to fuse features at corresponding anatomical locations across views. From the resulting features, FReSH jointly predicts scene depth, hand geometry, and camera parameters.
To support accurate 3D hand reconstruction across capture conditions, we pretrain the local hand branch on a large corpus that combines diverse single-view images with synthetic multi-view observations from randomized viewpoints.

To integrate evidence across time as well as across views, we extend the spatial model with a bounded causal memory that reuses hand and scene features from earlier observations. A learned registration module expresses successive reconstructions in the world frame established by the initial observation.

We evaluate FReSH on benchmarks for multi-view hand reconstruction and monocular hand--scene video reconstruction. FReSH achieves SoTA hand reconstruction from multi-view images, even \textbf{surpassing calibrated methods} in Procrustes-aligned accuracy. It also delivers competitive video reconstruction with more accurate scene depth than existing hand--scene methods. With multiple uncalibrated video streams, FReSH achieves further gains over monocular reconstruction, highlighting the value and potential of uncalibrated multi-view methods.

Our main contributions are as follows:
\begin{itemize}
    \item We propose FReSH, a feed-forward framework that jointly reconstructs hand motion, scene geometry, and camera trajectories in a shared world frame from sparse uncalibrated views, supporting scalable capture of human manipulation.

    \item We design global and local branches that combine scene context with detailed hand articulation through bidirectional feature exchange and anatomical grounding.

    \item We extend FReSH to multi-view videos through causal hand--scene memory and registration to a persistent world frame. Reusing bounded historical features supports memory- and runtime-efficient temporal reconstruction.

    \item Extensive experiments demonstrate state-of-the-art multi-view hand reconstruction and competitive hand–scene video reconstruction
\end{itemize}

\begin{figure*}[t]
    \centering
    \includegraphics[width=\linewidth]{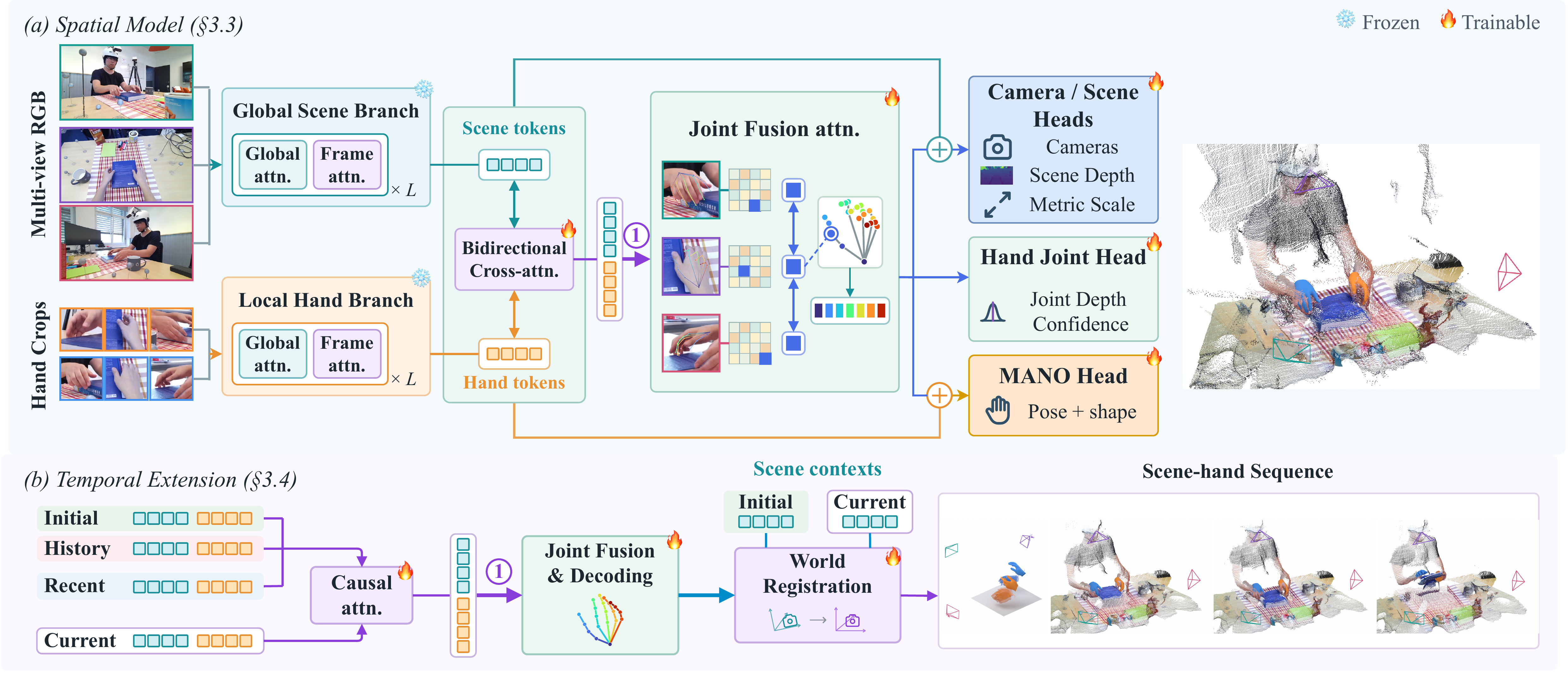}
    \vspace{-2em}
    \caption{\textbf{Pipeline of FReSH.} (a) Joint hand--scene reconstruction from uncalibrated views. (b) Causal token updates at \textcircled{1} and world registration extend the model to video.}
    \vspace{-1.5em}
    \label{fig:pipeline}
\end{figure*}

\section{Related Work}

\noindent\textbf{Hand Reconstruction} aims to recover 3D hand shape, pose, and motion from visual observations.
Single-view methods estimate 3D hand pose and shape by regressing MANO parameters~\cite{romero2022embodied,pavlakos2024reconstructing,potamias2025wilor} or directly predicting mesh vertices~\cite{choi2020pose2mesh,ge20193d,kulon2020weakly}.
Since one view leaves depth and occluded articulation ambiguous, multi-view methods combine complementary observations using known camera geometry~\cite{yang2023poem,yang2024POEM} or learn to fuse features without supplied camera extrinsics~\cite{ali2025handmvnet,chen2025a3net}.
Video methods use neighboring frames when occlusion, motion blur, or low resolution obscures hand details~\cite{fu2023deformer,fan2025temporalhand}.
HaPTIC~\cite{ye2026haptic} uses spatial context by letting hand features attend to full-frame features to recover camera-space hand trajectories from monocular video.
Recovering a world-space trajectory also requires disentangling hand motion from camera motion.
Staged pipelines combine hand estimates with independently tracked cameras, so camera trajectory errors can carry over to the recovered hand motion~\cite{zhang2025hawor,yu2025dynhamr}.
Concurrent models jointly estimate hand and camera motion from monocular egocentric video~\cite{zhu2026mint,suzuki2026dewhand}.
FReSH fuses corresponding anatomical joint features across uncalibrated views and exchanges hand--scene features bidirectionally to jointly estimate hands, cameras, and scene geometry.

\noindent\textbf{Joint Hand--Scene Reconstruction} seeks to recover hand motion and the geometry of the surrounding workspace in a shared world frame.
Hand--object methods capture immediate interactions by reconstructing articulated hands and the geometry of manipulated objects~\cite{fan2024hold,liu2025easyhoi,wang2024Magichoi}.
WHOLE~\cite{ye2026whole} uses object templates to recover hand and object motion in world coordinates from egocentric video.
Beyond individual objects, concurrent models jointly estimate hands, camera motion, and workspace geometry from monocular egocentric video~\cite{lyu2026reviv,fu2026egofound3r,ji2026egoxperience,pei2026egohsi3r}.
With a single camera, hidden articulation must be inferred from indirect evidence, whereas a synchronized second view can reveal otherwise occluded hand regions.
FReSH supports one or a few uncalibrated cameras, combines simultaneous observations when multiple views are available, and registers successive reconstructions to a common world frame.

\noindent\textbf{Multi-view 3D Reconstruction} recovers geometry and, optionally, appearance from images of an object or scene taken from different viewpoints.
Classical multi-view stereo uses photometric matching across calibrated views~\cite{galliani2015massively,schonberger2016pixelwise}, while learned variants aggregate evidence through cost volumes or point features, typically retaining known camera geometry to relate views~\cite{yao2018mvsnet,gu2020cascade,chen2019PointMVSNet}.
Feed-forward models relax calibration requirements by learning geometry directly from images, progressing from pairwise point-map prediction~\cite{dust3r_cvpr24,mast3r_eccv24} to joint camera and scene estimation across views, as in VGGT and VGGT\_O~\cite{wang2025vggt,wang2026vggtomega}.
Causal models such as StreamVGGT and LingBot-Map retain geometric context as new frames arrive~\cite{cut3r,zhuo2025streamvggt,chen2026lingbotmap}.
For hand--scene reconstruction, however, full-image scene encoders can lose fine articulation cues when fingers occupy only a few pixels at the input resolution.
FReSH exchanges features between full images and hand crops and retains both in its bounded causal memory.

\section{Method}
\label{sec:method}

As shown in~\autoref{fig:pipeline}, FReSH encodes full images and hand crops in separate branches, exchanges features between them, and predicts 2D joints in each view.
FReSH fuses features at corresponding joints across views before decoding cameras, scene depth, and hand geometry, then places each hand mesh in the scene using metric 3D joint anchors.

\subsection{Problem Formulation}
\label{sec:problem_definition}

Given full RGB images from one or more uncalibrated cameras and the associated hand detections~\cite{xu2022vitpose}, we aim to jointly reconstruct the 3D hands, scene geometry, and camera poses in a shared world frame. At time $t\in\{1,\ldots,T\}$, $\mathcal I_t=\{I_{t,v}\}_{v=1}^{V_t}$ denotes the synchronized images from $V_t$ views. Each hand detection provides a 2D bounding box and a handedness label. For unannotated video, we can derive both from the hand keypoints predicted by an off-the-shelf pose estimator such as ViTPose~\cite{xu2022vitpose}. For spatial reconstruction, we set $T=1$ and use only the images captured at that time step. The first camera in $\mathcal I_1$ defines the world frame for the entire sequence.

For each camera $v$, we estimate an intrinsic matrix $K_v$ and hold it fixed over time.
For each observation $(t,v)$, we estimate the camera-to-world pose $C_{t,v}=(R_{t,v},\mathbf c_{t,v})$ and a metric scene depth map $D_{t,v}$.
Here $R_{t,v}$ is the camera-to-world rotation and $\mathbf c_{t,v}$ is the camera center in world coordinates.
Back-projecting $D_{t,v}$ using $K_v$ and $C_{t,v}$ yields a point cloud of the visible scene in the world frame.
A positive scale factor $s$ converts the initial reconstruction from canonical to metric units.

For each hand $h$, we use the MANO model~\cite{romero2022embodied} to represent hand articulation pose $\boldsymbol\theta_t^h\in\mathbb R^{45}$ and hand shape with coefficients $\boldsymbol\beta^h\in\mathbb R^{10}$ shared across time. A global orientation $\boldsymbol\omega_t^h\in\mathbb R^3$ in axis-angle form and a translation $\mathbf r_t^h\in\mathbb R^3$ place the hand mesh in the world frame. The sequence $\{(\boldsymbol\theta_t^h,\boldsymbol\omega_t^h,\mathbf r_t^h)\}_{t=1}^T$ describes the hand motion.

\subsection{Hand Reconstruction Pretraining}
\label{sec:hand_pretraining}

\begin{figure}[t]
    \centering
    \includegraphics[width=0.85\columnwidth]{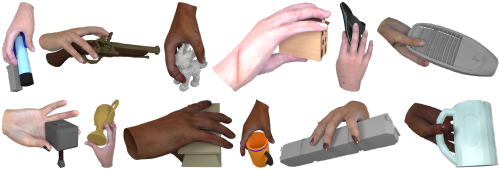}
    \caption{\textbf{Synthetic pretraining data.} Rendered hand--object examples span diverse objects, grasps, and hand appearances.}
    \vspace{-1em}
    \label{fig:pretraining_data}
\end{figure}

Since VGGT~\cite{wang2025vggt} has demonstrated the effectiveness of alternating within-view and cross-view attention for learning multi-view geometric representations, we adapt this aggregation design to our local hand branch. To support accurate and robust multi-view hand reconstruction across diverse inputs, we pretrain the branch on heterogeneous data from multiple sources, combining real single-view images~\cite{pavlakos2024reconstructing} with synthetic multi-view images, as shown in \autoref{fig:pretraining_data}, and using the annotations available in each source. Real single-view data provide diverse hand poses and image conditions, while synthetic multi-view hand--object groups provide geometric supervision under varied camera configurations. The training is supervised with 2D keypoint reprojection, supplemented by MANO parameters and wrist-relative 3D joints where annotated. The synthetic groups additionally supervise relative camera geometry and constrain camera predictions through reprojection of the shared 3D hand. Training on heterogeneous data requires balancing objectives with different loss formulations. We therefore alternate single-view and multi-view batches and accumulating gradients from both batch types at each update. The resulting weights initialize FReSH's local branch and MANO head, providing strong geometric priors for accurate and robust multi-view hand reconstruction.
More details are provided in the supplementary material.

\subsection{Spatial Reconstruction}
\label{sec:spatial_reconstruction}

Accurate hand--scene reconstruction requires both fine-grained hand details and global scene context. Processing full images at a fixed resolution can obscure finger articulation when hands occupy only a small image region. Resized hand crops preserve more detail but discard surrounding scene cues needed for camera estimation and global hand placement. FReSH therefore combines global scene reasoning with local hand reasoning to jointly recover detailed hand geometry and its spatial relationship to the scene.

\subsubsection{Global and Local Geometric Reasoning}
\label{sec:model_design}

We encode full images and hand crops in separate branches and exchange features before decoding, allowing global context and local hand details to jointly inform hand and camera reconstruction.

\noindent\textbf{Global scene branch.}
VGGT-$\Omega$~\cite{wang2026vggtomega} encodes the full images into scene tokens $G_t$. These tokens retain the scene structure and cross-view context needed to estimate camera poses and scene depth.

\noindent\textbf{Local hand branch.}
For each hand, the local encoder embeds its crops with DINOv2 ViT-L/14~\cite{oquab2023dinov2} and aggregates their features with alternating attention within and across views, following VGGT~\cite{wang2025vggt}. The crops give fine joint configurations more spatial resolution. The resulting crop tokens and learned MANO queries form the local representation $L_t$. Left and right hands form separate crop groups, and each crop retains its affine mapping to the full image. Section~\ref{sec:hand_pretraining} describes how we pretrain this branch.

\noindent\textbf{Bidirectional feature exchange.}
After projecting the two representations to a common feature space, a bidirectional block applies
\begin{equation}
\begin{aligned}
    G_t' &= G_t+\mathcal A_{G\leftarrow L}(G_t,L_t),\\
    L_t' &= L_t+\mathcal A_{L\leftarrow G}(L_t,G_t),
\end{aligned}
\label{eq:bidirectional}
\end{equation}
where $\mathcal A(Q,X)$ denotes cross-attention with queries $Q$ and keys and values from $X$. Both directions read the pre-update features. We omit normalization and feed-forward layers for clarity. Spatial positions and handedness accompany the visual features, and invalid observations are masked. The scene update carries hand detail into camera and depth prediction, and the hand update gives the joint queries access to full-image context.

\noindent\textbf{Per-view 2D joint prediction.}
Cross-view correspondence is central to multi-view reconstruction, and hand joints provide natural anatomical correspondences. With camera geometry and 3D hand pose unknown, we localize these joints using 21 learned joint queries shared across views. Each query uses hand and scene features to predict the joint's 2D location $\mathbf u_{v,j}$ in the cropped image and its initial confidence. 

\subsubsection{Grounding Articulation in Scene Geometry}
\label{sec:joint_grounding}

A joint's image location guides feature fusion, and its back-projected 3D position anchors the decoded hand in the scene. We omit time and hand indices below.

\noindent\textbf{Cross-view joint fusion.}
For each hand joint, we fuse crop features sampled around its predicted 2D location in each view, giving greater weight to observations with higher initial confidence, and conduct an attention operation among different joints within each view. The fused features update the scene tokens and hand queries before decoding. From the updated joint features, we predict canonical joint depth and re-estimate confidence for each joint.

\noindent\textbf{Decoding and hand placement.}
\label{sec:mano_head}
The camera and depth heads predict camera intrinsics, poses, and canonical scene depths from the updated scene tokens. The MANO head decodes hand pose and shape, which represent wrist-relative geometry expressed in the first camera's frame. A scale head uses pooled hand and scene features to predict one factor $s$ for converting camera translations, scene depths, and joint depths to metric units.

\begin{wrapfigure}{r}{0.3\columnwidth}
    \vspace{-\intextsep}
    \centering
    \includegraphics[width=\linewidth]{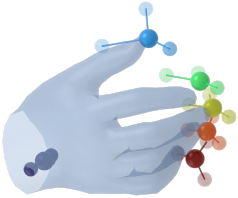}
    \caption{Anchor-guided hand placement.}
    \label{fig:geometric_grounding}
    \vspace{-\intextsep}
\end{wrapfigure}

As shown in \autoref{fig:geometric_grounding}, we back-project the 2D joints at their metric depths and robustly align the wrist-relative hand joints to these anchors using one translation $\mathbf r$ per hand. We give greater weight to anchors with higher confidence.

The same hand joints thus inform camera estimation and provide the anchors that place each articulated hand in the reconstructed scene. The supplementary material gives more details.

\subsection{Temporal Reconstruction}
\label{sec:temporal_memory}
Observations across time provide complementary evidence for refining hand--scene reconstruction. However, camera motion can change both visibility and the local reference frame between time steps. Therefore, we integrate current and past observations through causal attention to refine each reconstruction, then register the resulting geometry to the world frame.

\noindent\textbf{Bounded causal memory.}
To provide causal context in an efficient way, we maintain a bounded memory during temporal reconstruction~\cite{zhuo2025streamvggt,chen2026lingbotmap}.
This memory retains the initial view group as a persistent reference, recent groups for short-term context, and a sparse sample of older groups for longer-range evidence, keeping memory usage and per-step attention costs bounded as the video progresses.

We store projected keys and values from these groups in a memory bank $\mathcal M_{<t}$ and utilize a causal attention to update the scene-hand context. We use the tokens produced by the spatial model's bidirectional global--local attention as the current input $X_t$. At each causal attention layer, we concatenate the cached keys $K_{\mathcal M_{<t}}$ and values $V_{\mathcal M_{<t}}$ with their current counterparts and update the current tokens as follows.
\begin{equation}
\begin{aligned}
    X_t' &= X_t+\operatorname{Attn}(Q(X_t),\overline K_t,\overline V_t),\\
    \overline K_t&=[K_{\mathcal M_{<t}};K(X_t)],\quad
    \overline V_t=[V_{\mathcal M_{<t}};V(X_t)].
\end{aligned}
\label{eq:temporal_attention}
\end{equation}
Here $Q(\cdot)$, $K(\cdot)$, and $V(\cdot)$ are learned projections, and $[\cdot;\cdot]$ denotes token-wise concatenation. This residual update incorporates historical evidence into both global scene and local hand representations to refine the current reconstruction. With a bounded memory, cache storage and attention cost per step are independent of video length. 

\noindent\textbf{World registration.}
\label{sec:cam_head}
Each spatial reconstruction uses its current first camera as a local reference, which changes as the camera moves. We therefore register every group to the fixed world frame defined by the first camera at the initial time step. Using the current scene context and that retained from the initial group, a registration head predicts a rotation $R_t^a$, a translation $\mathbf a_t$, and a positive scale ratio $\alpha_t$. Together, they map the current canonical frame to the initial canonical frame.

Let $(R^0_{t,v},\mathbf c^0_{t,v})$ denote the local camera rotation and center, $D^0_{t,v}$ the canonical scene depth map, and $z^0_{t,v,j}$ the canonical depth of hand joint $j$ in view $v$. The superscript $0$ denotes predictions before registration. Combining the predicted transform with the initial canonical-to-metric scale $s$ gives
\begin{equation}
\begin{aligned}
    &R_{t,v} = R_t^aR^0_{t,v}, \qquad
    \mathbf c_{t,v} = s\left(\alpha_tR_t^a\mathbf c^0_{t,v}
        +\mathbf a_t\right),\\
    &D_{t,v} = s\alpha_tD^0_{t,v},\qquad
    z_{t,v,j}=s\alpha_tz^0_{t,v,j}.
\end{aligned}
\label{eq:registration}
\end{equation}
The first two expressions give each camera's orientation and metric position in the fixed world frame. The scene depths $D_{t,v}$ and joint depths $z_{t,v,j}$ remain measured in each camera's own frame, so they require only the shared scale conversion $s\alpha_t$. Back-projecting these depths with the registered cameras yields world-space scene points and joint anchors for hand placement. Hand and camera motion can thus be expressed in a common frame throughout the sequence. At initialization, $R_1^a=I$, $\mathbf a_1=\mathbf 0$, and $\alpha_1=1$.

We retain the initial metric scale and each camera's intrinsics, and fix each hand's shape at its first valid observation. Later groups may select or reorder cameras from this initialized pool while their extrinsics change. These quantities and the world reference persist across input chunks.

\subsection{Training Strategy}
\label{sec:training_strategy}

To learn accurate spatial geometry before incorporating temporal context, we train FReSH in two stages. We first train the spatial model to jointly reconstruct hands, scenes, and cameras from individual groups of synchronized views. During this stage, both pretrained image encoders remain frozen, while the joint fusion modules and prediction heads are trained. We then train the causal attention and world registration modules on video sequences, while fine-tuning the trainable components of the spatial model at a lower learning rate.

Both stages supervise the predicted hands, cameras, and scene geometry wherever annotations are available. We also align the reconstructed hand joints with the predicted 3D anchors and reproject them through the predicted cameras. The reprojection loss encourages the hand and camera predictions to fit the same 2D observations. The supplementary material details the losses and training protocol.

\begin{table*}[htp]
\centering
\footnotesize
\caption{\textbf{Multi-view hand reconstruction on DexYCB-Mv and HO3D-Mv.} We compare calibrated and calibration-free methods using root-relative and Procrustes-aligned joint and vertex metrics. \bestlegend{Green}/\secondlegend{blue} mark the best/second-best calibration-free results; gray rows denote methods using GT cameras.}
\vspace{-1em}
\label{tab:standard_hand}
\renewcommand{\arraystretch}{1.15}
\setlength{\dashlinedash}{0.6pt}
\setlength{\dashlinegap}{1.8pt}
\setlength{\tabcolsep}{3pt}

\resizebox{\linewidth}{!}{%
\begin{tabular}{lllc*{8}{c}}
\toprule
\multirow{2}{*}{Dataset} & \multirow{2}{*}{\#} & \multirow{2}{*}{Method} & \multirow{2}{*}{Cali-Free}
& \multicolumn{4}{c}{Root Relative} & \multicolumn{4}{c}{Procrustes Alignment}\\
\cmidrule(lr){5-8}\cmidrule(lr){9-12}
& & & & MPVPE$\downarrow$ & AUC$_{V@20}\uparrow$ & MPJPE$\downarrow$ & AUC$_{J@20}\uparrow$ & MPVPE$\downarrow$ & AUC$_{V@20}\uparrow$ & MPJPE$\downarrow$ & AUC$_{J@20}\uparrow$\\
\midrule
\multirow{7}{*}{\shortstack[l]{DexYCB-Mv\\(8 views)}}
 & \texttt{1} & {\color{graytext}\POEMs$^{\S}$~\cite{yang2023poem}} & \xmark & \color{graytext}7.21 & \color{graytext}0.65 & \color{graytext}7.30 & \color{graytext}0.62 & \color{graytext}4.00 & \color{graytext}0.80 & \color{graytext}3.93 & \color{graytext}0.78\\
 & \texttt{2} & {\color{graytext}\POEMg-large$^{\ast}$~\cite{yang2024POEM}} & \xmark & \color{graytext}7.50 & \color{graytext}0.64 & \color{graytext}7.54 & \color{graytext}0.63 & \color{graytext}3.84 & \color{graytext}0.81 & \color{graytext}3.68 & \color{graytext}0.81\\
\cdashline{3-12}
 & \texttt{4} & HandMvNet w/o cam.$^{R}$~\cite{ali2025handmvnet} & \cmark & 8.52 & \secondresult{0.63} & \secondresult{8.42} & \secondresult{0.64} & 4.54 & 0.77 & 4.15 & 0.79\\
 & \texttt{5} & HandMvNet-HR w/o cam.$^{R}$~\cite{ali2025handmvnet} & \cmark & \secondresult{8.04} & \bestresult{0.64} & \bestresult{7.79} & \bestresult{0.66} & 4.42 & 0.78 & \secondresult{4.13} & \secondresult{0.80}\\
 & \texttt{3} & VGGT-$\Omega$+POEM~\cite{wang2026vggtomega,yang2024POEM} & \cmark & 9.93 & 0.52 & 11.34 & 0.45 & 4.31 & 0.79 & 5.11 & 0.75\\
 & \texttt{3a} & Ours-cam+POEM & \cmark & \bestresult{7.73} & 0.62 & 8.69 & 0.58 & \secondresult{3.98} & \secondresult{0.80} & 5.03 & 0.75\\
 & \texttt{6} & Ours-spatial & \cmark & 9.41 & 0.56 & 9.73 & 0.55 & \bestresult{3.75} & \bestresult{0.81} & \bestresult{3.86} & \bestresult{0.81}\\
\midrule
& & & & & AUC$_{V@50}\uparrow$ & & AUC$_{J@50}\uparrow$ & & AUC$_{V@50}\uparrow$ & & AUC$_{J@50}\uparrow$\\
\multirow{7}{*}{\shortstack[l]{HO3D-Mv\\(5 views)}}
 & \texttt{11} & {\color{graytext}\POEMs$^{\S}$~\cite{yang2023poem}} & \xmark & \color{graytext}21.45 & \color{graytext} 0.58 & \color{graytext}21.94 & \color{graytext} 0.55 & \color{graytext}9.97 & \color{graytext} 0.80 & \color{graytext}9.60 & \color{graytext} 0.78\\
 & \texttt{12} & {\color{graytext}\POEMg-large$^{\ast}$~\cite{yang2024POEM}} & \xmark & \color{graytext}10.58 & \color{graytext}0.79 & \color{graytext}10.58 & \color{graytext} 0.79 & \color{graytext}4.59 & \color{graytext}0.91 & \color{graytext}4.13 & \color{graytext}0.91\\
\cdashline{3-12}
 & \texttt{14} & HandMvNet w/o cam.$^{R}$~\cite{ali2025handmvnet} & \cmark & 19.81 & 0.61 & 20.93 & 0.59 & 9.40 & 0.81 & 9.99 & 0.80\\
 & \texttt{15} & HandMvNet-HR w/o cam.$^{R}$~\cite{ali2025handmvnet} & \cmark & 23.17 & 0.56 & 24.19 & 0.55 & 10.93 & 0.78 & 11.22 & 0.78\\
 & \texttt{13} & VGGT-$\Omega$+POEM~\cite{wang2026vggtomega,yang2024POEM} & \cmark & \secondresult{11.41} & \secondresult{0.77} & 12.53 & 0.75 & 5.63 & 0.89 & 5.85 & 0.88\\
 & \texttt{13a} & Ours-cam+POEM & \cmark & \bestresult{9.95} & \bestresult{0.80} & \bestresult{10.58} & \bestresult{0.79} & \secondresult{4.65} & \secondresult{0.91} & \secondresult{5.14} & \secondresult{0.90}\\
 & \texttt{16} & Ours-spatial & \cmark & 11.65 & \secondresult{0.77} & \secondresult{12.08} & \secondresult{0.76} & \bestresult{4.21} & \bestresult{0.92} & \bestresult{4.34} & \bestresult{0.91}\\
\bottomrule
\end{tabular}}
\vspace{-1.5em}
\end{table*}

\section{Experiments}
\label{sec:experiments}

We evaluate hand articulation, motion, and scene geometry to distinguish accurate hand shape from accurate placement in a shared frame. We then examine the benefit of additional views and the role of local--global interaction.

\noindent\textbf{Training data.}
We pretrained the hand branch on single-view images from the HaMeR training collection~\cite{pavlakos2024reconstructing} and synthetic multi-view hand--object groups. We generated the synthetic groups by rendering GraspXL grasps~\cite{zhang2024graspxl} on Objaverse objects~\cite{deitke2023objaverse} from randomized viewpoints, using DART hand textures~\cite{gao2022dart}. We trained Ours-spatial on synchronized views from HO3D~\cite{hampali2020honnotate}, DexYCB~\cite{chao2021dexycb}, H2O~\cite{kwon2021h2o}, GigaHands~\cite{fu2025gigahands}, and HOT3D~\cite{banerjee2025hot3d}. For temporal training, we used video sequences from HO3D, DexYCB, H2O, and HOT3D with one to five input views. The supplementary material gives the dataset sizes and sampling settings.

\noindent\textbf{Compute and runtime.}
Training required about 2,500 A100 GPU-hours. On an RTX 4090, a five-view HO3D forward pass takes 0.45\,s; a 501-step five-view H2O sequence takes about 8 minutes, including model loading and export.

\begin{table}[t]
\centering\footnotesize
\setlength{\tabcolsep}{2.5pt}
\renewcommand{\arraystretch}{1.15}
\caption{\textbf{Monocular video hand reconstruction on DexYCB across occlusion levels.} We report Procrustes-aligned joint error (PA-J) and PCK AUC over all frames and the 50--75\% and 75--100\% occlusion ranges. \bestlegend{Green}/\secondlegend{blue} mark the best/second-best results.}
\vspace{-1em}
\label{tab:dexycb_temporal}
\begin{tabular}{@{}lrrrrrr@{}}
\toprule
\multirow{2}{*}{Method} & \multicolumn{2}{c}{All} & \multicolumn{2}{c}{50\%--75\%} & \multicolumn{2}{c}{75\%--100\%}\\
\cmidrule(lr){2-3}\cmidrule(lr){4-5}\cmidrule(lr){6-7}
& PA-J$\downarrow$ & AUC$\uparrow$ & PA-J$\downarrow$ & AUC$\uparrow$ & PA-J$\downarrow$ & AUC$\uparrow$\\
\midrule
$S^{2}$HAND(V)~\cite{tu2023consistent} & 7.27 & 85.5 & 7.71 & 84.6 & 7.87 & 84.3\\
VIBE~\cite{kocabas2020vibe} & 6.43 & 87.1 & 6.84 & 86.4 & 7.06 & 85.8\\
TCMR~\cite{choi2021tcmr} & 6.28 & 87.5 & 6.58 & 86.8 & 6.95 & 86.1\\
Deformer~\cite{fu2023deformer} & 5.22 & 89.6 & 5.70 & 88.6 & 6.34 & 87.3\\
HaWoR~\cite{zhang2025hawor} & \secondresult{4.76} & \secondresult{90.5} & \secondresult{5.03} & \secondresult{89.9} & \secondresult{5.07} & \bestresult{89.9}\\
\midrule
Ours & \bestresult{4.30} & \bestresult{91.2} & \bestresult{4.59} & \bestresult{90.1} & \bestresult{5.05} & \bestresult{89.9}\\
\bottomrule
\end{tabular}
\vspace{-1em}
\end{table}

\begin{table*}[t]
\centering\footnotesize
\setlength{\tabcolsep}{3pt}
\renewcommand{\arraystretch}{1.1}
\caption{\textbf{Joint hand--scene video reconstruction on H2O and HOT3D.} We compare monocular methods with FReSH using one or all available synchronized views ($V$), evaluating aligned and world-space hand reconstruction together with cameras and depth. \bestlegend{Green}/\secondlegend{blue} mark the best/second-best monocular results per dataset. Units are mm for hand and C-T errors, degrees for C-R, and cm for D-MAE. $\dagger$: HaWoR's HOT3D scores cover 346/349 clips because SLAM fails on three; N/A denotes an unavailable output or GT.}
\vspace{-1em}
\label{tab:video_reconstruction}
\label{tab:standard_video}
\label{tab:joint_scene}
\resizebox{0.93\linewidth}{!}{%
\begin{tabular}{llc*{9}{r}}
\toprule
\multirow{2}{*}{Dataset} & \multirow{2}{*}{Method} & \multirow{2}{*}{$V$}
& \multicolumn{4}{c}{Aligned and root-relative hand errors}
& \multicolumn{5}{c}{Scene-space hands, cameras, and depth}\\
\cmidrule(lr){4-7}\cmidrule(lr){8-12}
& & & \shortstack{PA-\\MPJPE$\downarrow$} & \shortstack{WA-\\MPJPE$\downarrow$} & \shortstack{W-\\MPJPE$\downarrow$} & \shortstack{RR-\\MPJPE$\downarrow$}
& \shortstack{Abs-\\MPJPE$\downarrow$} & \shortstack{Abs-\\MPVPE$\downarrow$} & C-R$\downarrow$ & C-T$\downarrow$ & D-MAE$\downarrow$\\
\midrule
\multirow{5}{*}{H2O} & HaWoR~\cite{zhang2025hawor} & 1 & \secondresult{6.45} & 20.09 & 60.33 & \bestresult{14.12} & 98.95 & 99.07 & \bestresult{0.40} & 70.15 & N/A\\
 & ReViV~\cite{lyu2026reviv} & 1 & 10.81 & 25.37 & 62.89 & 28.45 & 73.31 & N/A & 4.02 & 47.19 & 113.95\\
 & EgoFound3R~\cite{fu2026egofound3r} & 1 & 6.98 & \bestresult{14.62} & \bestresult{32.09} & \secondresult{14.41} & \bestresult{41.29} & \bestresult{41.43} & \secondresult{1.56} & \secondresult{36.02} & \secondresult{10.43}\\
 & FReSH & 1 & \bestresult{6.04} & \secondresult{17.95} & \secondresult{42.81} & 15.70 & \secondresult{59.73} & \secondresult{59.53} & 4.28 & \bestresult{32.07} & \bestresult{6.16}\\
\cmidrule(lr){2-12}
 & FReSH & 5 & 4.73 & 14.52 & 33.21 & 14.53 & 49.81 & 49.50 & 4.20 & 31.59 & 3.11\\
\midrule
\multirow{5}{*}{HOT3D Aria} & HaWoR~\cite{zhang2025hawor}$^{\dagger}$ & 1 & \bestresult{7.98} & \bestresult{79.21} & 325.79 & \bestresult{41.92} & 802.35 & 803.08 & 12.95 & 290.23 & N/A\\
 & ReViV~\cite{lyu2026reviv} & 1 & 19.83 & 122.33 & 376.63 & 109.43 & 576.86 & N/A & \bestresult{5.55} & \secondresult{96.14} & N/A\\
 & EgoFound3R~\cite{fu2026egofound3r} & 1 & 19.30 & 94.44 & \secondresult{273.13} & 87.71 & \secondresult{328.86} & \secondresult{329.05} & \secondresult{6.43} & \bestresult{76.97} & N/A\\
 & FReSH & 1 & \secondresult{8.00} & \secondresult{94.33} & \bestresult{207.73} & \secondresult{44.66} & \bestresult{207.89} & \bestresult{206.51} & 23.65 & 148.24 & N/A\\
\cmidrule(lr){2-12}
 & FReSH & 3 & 7.83 & 95.27 & 209.48 & 44.44 & 206.73 & 205.39 & 23.67 & 148.18 & N/A\\
\bottomrule
\end{tabular}}
\vspace{-1em}
\end{table*}

\begin{figure*}[hpt]
  \centering
  \begin{minipage}[t]{0.7492\textwidth}
    \vspace{0pt}
    \centering
    \captionsetup{hypcap=false}
  \includegraphics[width=\linewidth]{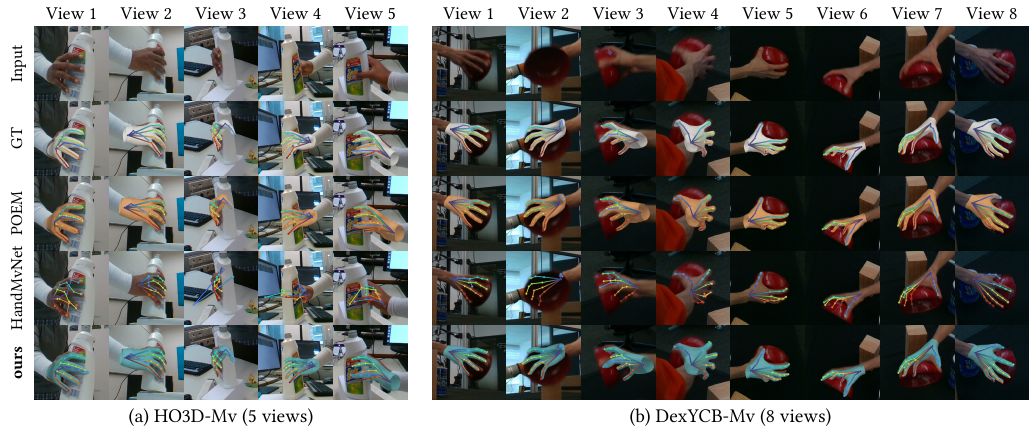}
  \vspace{-2em}
  \caption{\textbf{Multi-view hand reconstruction.}
  Selected HO3D-Mv (left) and DexYCB-Mv (right) examples use matched crops.
  HandMvNet shows native 2D joints.}
  \vspace{-1em}
  \label{fig:qual_spatial}
  \end{minipage}\hfill
  \begin{minipage}[t]{0.2308\textwidth}
    \vspace*{9.14bp}
    \centering
    \captionsetup{hypcap=false}
    \input{figures/qualitative_overlay}
  \end{minipage}
\vspace{-1em}
\end{figure*}

\subsection{Evaluation Metrics}
\label{sec:evaluation_metrics}

\noindent\textbf{Hand geometry.}
MPJPE and MPVPE measure mean joint and vertex errors in millimeters. Root-relative (RR) scores remove wrist translation; Procrustes-aligned (PA) scores also remove global rotation and scale. AUC is the normalized area under the PCK curve over each table's thresholds, reported in $[0,1]$ for spatial reconstruction and as a percentage over 0--50\,mm for DexYCB video. DexYCB video uses camera-frame PA-MPJPE; branch-ablation hand scores use each hand's reference-camera frame.

\noindent\textbf{Temporal alignment.}
For H2O and HOT3D right-hand motion, we use HaWoR's similarity-fitting operators~\cite{zhang2025hawor} with windows of up to 150 consecutive valid observations. PA-MPJPE aligns each frame independently; WA-MPJPE fits the entire window; W-MPJPE fits its first two frames and applies the transform throughout.

\noindent\textbf{Global placement and scene geometry.}
Abs-MPJPE and Abs-MPVPE measure both hands' errors in the fixed first-camera world frame without GT pose or scale fitting. C-R, C-T, and D-MAE measure camera rotation (degrees), camera-center (mm), and mean absolute depth (cm) errors, respectively. Spatial camera averages exclude the reference view. Branch-ablation P-Err measures corresponding-pixel world-point error (cm). Detailed alignment and windowing conventions are provided in the supplementary material.

\subsection{Reconstruction from Multi-view Images}
\label{sec:training_data}

\noindent\textbf{Setup.}
We evaluate our spatial model on synchronized views from individual time steps in DexYCB~\cite{chao2021dexycb} and HO3D~\cite{hampali2020honnotate}. Additional results on OakInk~\cite{yang2022oakink}, InterHand~\cite{Moon_2020_ECCV_InterHand26M}, and ARCTIC~\cite{fan2023arctic} are provided in the supplementary material.

\noindent\textbf{Baselines.}
HandMvNet~\cite{ali2025handmvnet} predicts root-relative hand joints and meshes from multiple views without estimating cameras; we evaluate its standard and high-resolution variants without camera calibration. POEM~\cite{yang2023poem,yang2024POEM} reconstructs hand meshes from calibrated views and provides the GT-camera reference. Both operate on hand crops, whereas FReSH also uses full images for scene reconstruction. For uncalibrated input, VGGT-$\Omega$+POEM and Ours-cam+POEM supply cameras predicted by VGGT\_O~\cite{wang2026vggtomega} and FReSH, respectively, to the same frozen POEM-v2-large model, using the same GT-derived hand boxes as FReSH.

\noindent\textbf{Results.}
As shown in~\autoref{tab:standard_hand}, FReSH outperforms the uncalibrated baselines in aligned joint and mesh accuracy on both benchmarks, and even surpasses the calibrated references in aligned mesh reconstruction. The advantage is strongest in local hand geometry, highlighting its ability to recover detailed articulation without camera calibration.

\noindent\textbf{Qualitative Comparison.}
\autoref{fig:qual_spatial} presents multi-view hand reconstruction results on HO3D and DexYCB, with hand predictions projected onto synchronized views. FReSH captures the finger arrangement around manipulated objects and preserves consistent grasp articulation across viewpoints, supporting the quantitative gains in local hand geometry.

\subsection{Reconstruction from Video}

\noindent\textbf{Setup.}
We evaluate monocular hand reconstruction under different occlusion levels on DexYCB, and joint hand--scene reconstruction on H2O~\cite{kwon2021h2o} recordings (444--1,025 frames each) and HOT3D Aria~\cite{banerjee2025hot3d} clips (150 frames each). To assess additional views, we augment the ego RGB stream with four external RGB streams on H2O and two grayscale tracking streams on HOT3D. The supplementary material details the evaluation protocols.

\noindent\textbf{Baselines.}
For DexYCB, we include the temporal hand reconstruction methods $S^{2}$HAND(V)~\cite{tu2023consistent} and Deformer~\cite{fu2023deformer}, and hand-adapted versions of VIBE~\cite{kocabas2020vibe} and TCMR~\cite{choi2021tcmr}, originally developed for body reconstruction. We use the DexYCB baseline scores reported by HaWoR~\cite{zhang2025hawor}. For world-space reconstruction, HaWoR combines hand-mesh estimation with SLAM to recover hand and camera motion; scene depth is not a joint output. ReViV~\cite{lyu2026reviv} models wearer motion and scene geometry, while EgoFound3R~\cite{fu2026egofound3r} focuses on joint hand--scene reconstruction. FReSH and HaWoR share GT-derived 2D hand boxes, whereas ReViV and EgoFound3R retain their RGB-only interfaces. The world-space baselines use a single egocentric stream; FReSH also accepts synchronized uncalibrated views. HaWoR's HOT3D scores are marked as a successful subset because its SLAM pipeline failed on three clips.

\noindent\textbf{Monocular temporal hand reconstruction.}
\autoref{tab:dexycb_temporal} shows that FReSH achieves the lowest aligned joint error across all occlusion groups on DexYCB. Its advantage persists under heavy occlusion, demonstrating robust hand articulation recovery from monocular video.

\noindent\textbf{Hand articulation and motion.}
The monocular comparison in~\autoref{tab:video_reconstruction} shows that FReSH achieves leading camera-frame articulation accuracy (PA-MPJPE) on H2O and the lowest world-space motion error (W-MPJPE) on HOT3D. Its competitive performance across the remaining motion metrics shows that feed-forward reconstruction can recover both detailed hand articulation and motion in a shared world frame.

\begin{figure}[t]
  \centering
  \includegraphics[width=\linewidth]{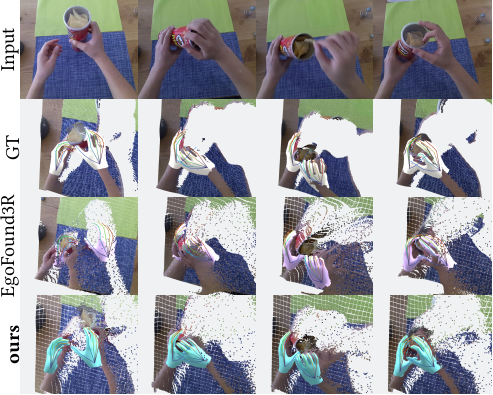}
  \caption{\textbf{Scene--hand reconstruction on H2O.}
  Four timestamps compare GT, EgoFound3R, and monocular FReSH from a fixed observer.}
  \vspace{-1em}
  \label{fig:qual_world}
\end{figure}

\noindent\textbf{Reconstruction in a shared frame.}
\autoref{tab:video_reconstruction} further demonstrates that the benefits extend to scene geometry and absolute hand placement. In the monocular comparison, FReSH provides the most accurate camera translations and depth on H2O, and the lowest absolute hand reconstruction errors on HOT3D. These results demonstrate its ability to recover hands and their surroundings in a common metric frame.

\noindent\textbf{Additional views.}
As shown in~\autoref{tab:video_reconstruction}, additional uncalibrated views substantially improve hand and scene reconstruction on H2O. On HOT3D, the gains are concentrated in hand articulation, with mixed changes in global motion. The same model accommodates these different camera configurations without retraining, highlighting the flexibility of our method.

\noindent\textbf{Qualitative Comparison.}
\autoref{fig:qual_overlay} presents monocular video hand reconstruction on H2O through overlays at successive timestamps, while~\autoref{fig:qual_world} shows joint hand--scene reconstruction from a fixed world-space viewpoint. FReSH accurately reconstructs fine-grained finger articulation and recovers grasp structure together with the surrounding geometry in a common coordinate frame. Predictions are shown without GT alignment, using each method's native projection convention.

\subsection{Ablation Studies}

\begin{table}[t]
\centering\footnotesize
\setlength{\tabcolsep}{2.2pt}
\renewcommand{\arraystretch}{1.15}
\caption{\textbf{Ablation of local--global interaction on H2O.} We isolate the local and global branches at inference using the same trained checkpoint, without retraining. \cmark/\xmark indicate enabled/disabled branches, and bold marks the complete coupled model. Units are mm for hand and C-T errors, degrees for C-R, and cm for D-MAE/P-Err; N/A denotes an unavailable output.}
\vspace{-1em}
\label{tab:branch_isolation}
\begin{tabular}{@{}ccrrrrrr@{}}
\toprule
\multirow{2}{*}{Local} & \multirow{2}{*}{Global}
& \multicolumn{2}{c}{Hand} & \multicolumn{2}{c}{Camera} & \multicolumn{2}{c}{Scene}\\
\cmidrule(lr){3-4}\cmidrule(lr){5-6}\cmidrule(lr){7-8}
& & RR-J$\downarrow$ & PA-J$\downarrow$ & C-R$\downarrow$ & C-T$\downarrow$ & D-MAE$\downarrow$ & P-Err$\downarrow$\\
\midrule
\cmark & \xmark & 14.06 & 4.76 & N/A & N/A & N/A & N/A\\
\xmark & \cmark & N/A & N/A & 4.62 & 131.73 & 20.85 & 23.49\\
\midrule
\cmark & \cmark & \textbf{10.15} & \textbf{3.88} & \textbf{1.74} & \textbf{48.61} & \textbf{3.29} & \textbf{4.87}\\
\bottomrule
\end{tabular}
\vspace{-1em}
\end{table}

\noindent\textbf{Local and global branches.}
\autoref{tab:branch_isolation} shows that local--global interaction improves all evaluated hand, camera, and scene metrics. We compare coupled and isolated inference from the same validation-selected checkpoint on the five-view H2O test set, keeping inputs fixed and disabling cross-branch interactions without retraining. The consistent gains demonstrate the benefit of combining fine hand articulation with global scene context for joint reconstruction. More ablation studies are detailed in the supplementary material.

\section{Conclusion}
We presented FReSH, a feed-forward framework that jointly reconstructs hands, scenes, and cameras from sparse uncalibrated videos. Its global--local architecture couples scene context with detailed hand observations to recover fine articulation, while causal memory and registration extend reconstruction across frames in a shared world. FReSH achieves lower Procrustes-aligned vertex errors than calibrated baselines, demonstrating accurate hand mesh reconstruction without camera calibration.

\noindent\textbf{Limitations and future work.}
Camera trajectory estimation on egocentric videos remains a main limitation. This likely stems in part from imbalanced training exposure: views are sampled uniformly, but static-camera views are more numerous. Future work will balance training exposure across camera types and model camera mode.

{
    \small
    \bibliographystyle{ieeenat_fullname}
    \bibliography{main}
}

\end{document}